\documentclass{article}

\usepackage{microtype}
\usepackage{graphicx}
\usepackage{subfigure}
\usepackage{booktabs} 
\usepackage{multirow}
\usepackage{amssymb}
\usepackage{pifont}
\usepackage{sidecap}
\usepackage[table]{xcolor}
\usepackage{amsmath}
\usepackage{amssymb}
\usepackage{float}

\usepackage[accepted]{icml2019} 
\usepackage{hyperref}

\newcommand{\norm}[1]{\left\lVert #1 \right\rVert}

\icmltitlerunning{Towards an Adversarially Robust Normalization Approach}

\begin{document}

\twocolumn[
\icmltitle{Towards an Adversarially Robust Normalization Approach}



\icmlsetsymbol{equal}{*}

\begin{icmlauthorlist}
\icmlauthor{Muhammad Awais}{khu}
\icmlauthor{Fahad Shamshad}{itu}
\icmlauthor{Sung-Ho Bae}{khu}
\end{icmlauthorlist}

\icmlaffiliation{khu}{Kyung-He University, South Korea}
\icmlaffiliation{itu}{Information Technology University, Lahore, Pakistan}

\icmlcorrespondingauthor{Muhammad Awais}{awais@khu.ac.kr}

\icmlkeywords{BatchNormalization, Deep Learning}

\vskip 0.3in
]
\setlength{\abovedisplayskip}{3pt}
\setlength{\belowdisplayskip}{3pt}


\printAffiliationsAndNotice{}  

\begin{abstract}
Batch Normalization (BatchNorm) is effective for improving the performance and accelerating the training of deep neural networks. However, it has also shown to be a cause of adversarial vulnerability, i.e., networks without it are more robust to adversarial attacks. In this paper, we investigate how BatchNorm causes this vulnerability and proposed new normalization that is robust to adversarial attacks. We first observe that adversarial images tend to shift the distribution of BatchNorm input, and this shift makes train-time estimated population statistics inaccurate. We hypothesize that these inaccurate statistics make models with BatchNorm more vulnerable to adversarial attacks. We prove our hypothesis by replacing train-time estimated statistics with statistics calculated from the inference-time batch. We found that the adversarial vulnerability of BatchNorm disappears if we use these statistics. However, without estimated batch statistics, we can not use BatchNorm in the practice if large batches of input are not available. To mitigate this, we propose Robust Normalization (RobustNorm); an adversarially robust version of BatchNorm. We experimentally show that models trained with RobustNorm perform better in adversarial settings while retaining all the benefits of BatchNorm. 
Code is available at \url{https://github.com/awaisrauf/RobustNorm}.
\end{abstract}

\section{Introduction}

Deep neural networks have shown impressive performance for image classification tasks. However, they are highly vulnerable to adversarial inputs; addition of small but targeted noise -- imperceptible to humans but make networks misclassify with high confidence \citep{goodfellow2014explaining, nguyen2015deep, carlini2017towards}. This vulnerability has severe consequences as these neural networks are employed in many security-critical applications such as face recognition, disease detection, and fraud prevention. Understanding the adversarial phenomenon is of high interest and many explanations such as blind spots in learned class boundaries, local linear nature of networks, and shift in input distribution have been presented \cite{szegedy2013intriguing, goodfellow2014explaining, ding2019sensitivity, ilyas2019adversarial}. Similarly, many defense mechanisms to prevent this vulerability have also emerged. There are two different types of defense mechanisms: adding adversarial examples for training also called adversarial training \cite{madry2017towards, tramer2017ensemble} and improving the architecture of neural networks \citep{ papernot2016distillation, xie2019feature}. 

While there exist a plethora of reasons for the adversarial behavior of neural networks \citep{jacobsen2018excessive, simon2018adversarial, yuan2019adversarial, ilyas2019adversarial, geirhos2018imagenet}, a recent study by \citealt{galloway2019batch} has shown that BatchNorm is one of them. They have empirically shown that we can enhance the robustness of neural networks against adversarial perturbations by removing BatchNorm. We first investigated how BatchNorm makes neural networks more vulnerable to adversarial attacks and provided a fresh perspective. Secondly, we propose a new normalization which is more robust than BatchNorm while also keeps the benefits of BatchNorm.


BatchNorm estimates population statistics (mean and variance) during training and uses them for inference. We hypothesize that the distribution shift in adversarial inputs makes these statistics inaccurate as they are estimated from clean images. To test our hypothesis, we replaced these train-time estimated statistics with current input batch statistics and showed improvement in adversarial robustness. However, we can not use this as a remedy since BatchNorm requires large batch size to calculate these statistics reliably (which may not be available at test time). Similarly, if we remove BatchNorm, we lose many benefits such as higher learning rate, faster convergence, and significant improvement in the accuracy, etc. \cite{hoffer2018norm}. To mitigate this, we propose a new normalization which is based on our insights and pervious work on understanding BatchNorm. This proposed normalization outperforms BatchNorm for adversarial accuracy on several datasets while keeping other benefits of it. We call this normalization RobustNorm for its robust properties.

\section{Related Work}
Previous works have provided many explanations to understand the adversarial vulnerability of neural networks. \citealt{szegedy2013intriguing} linked adversarial vulnerability to blind spots in the discontinuous classification boundary of the neural network, \citealt{goodfellow2014explaining} blamed it on the local linearity of neural networks and showed it by constructing an attack that leverages this property. Some recent work has connected it with random noise \cite{fawzi2016robustness, ford2019adversarial}, spurious correlations learned by neural networks \cite{ilyas2019adversarial}, insufficient data \cite{schmidt2018adversarially} high dimensions of input data \cite{gilmer2018adversarial, fawzi2018adversarial}, and distributional shift \cite{jacobsen2018excessive, ding2019sensitivity}. Similarly, researchers have also focused on constructing techniques to fight against these attacks. We can divide defense techniques into two major camps: training based defense in which we augment adversarial images with clean images during training \cite{madry2017towards, tramer2017ensemble} and architecture based defense \citep{ papernot2016distillation, xie2019feature} in which we change architecture base defense in which we change the network. Our work is related to the second camp.
    
Recently, \cite{galloway2019batch} empirically showed that the accelerated training properties and occasionally higher clean test accuracy of employing BatchNorm in the network come at the cost of low robustness to adversarial perturbations. They attributed it to the tilting of the decision boundary. Our work builds on their observation, but we gave a different perspective for understanding this behavior. Similarly, we also presented a new normalization to improve adversarial accuracy. Augmentation of adversarial images often results in lower clean accuracy. However, \cite{xie2019adversarial} got state of the art results on many classification datasets because they used different BatchNorm layer for clean and adversarial images during training. This shows the effect of the distributional shift introduced by adversarial examples on BatchNorm's statistic estimation. \emph{Note that, this work is closely related to our hypothesis.} 
    
Since the inception of BatchNorm, many different variants of it have been proposed and each variant of it solves a particular problem of original formulation. LayerNorm \cite{ba2016layer} solves the problem of fix batch size training making it useful in sequence models, BatchReNorm \cite{ioffe2017batch} and GroupNorm \cite{wu2018group} eases the problem of small-batch training making it functional for tasks like detection or segmentation and InstanceNorm \cite{ulyanov2016instance} reduces intra-batch dependency making it applicable in style transfer. Our work is related to these papers as we also propose a variant that solves the problem of adversarial vulnerability in BatchNorm. Our work is also dependent on some recent work on understanding BatchNorm. 

\section{Background}
We consider a standard classification task for data $\boldsymbol{x} \in \mathbb{R}^{n}$ and corresponding true labels $y \in \lbrace 1,2,...,k\rbrace$, sampled from a joint distribution $\mathbb{P}_{clean}(\boldsymbol{x},y)$. For learning, we divid this data into training ($\boldsymbol{x}_{t}, y_{t}$) and validation set ($\boldsymbol{x}_{v},y_{v}$). We denote deep neural network as a function, $\mathcal{F}_{\boldsymbol{\theta}}: \boldsymbol{x} \mapsto y$, where $\boldsymbol{\theta}$ denotes trainable parameters of the DNN. The parameter $\boldsymbol{\theta}$ is learned by minimizing a loss function $\mathcal{L}(\mathcal{F}_{\boldsymbol{\theta}}(\boldsymbol{x}_{t}), y_{t})$. We denote clean accuracy of a neural network as percentage of $\boldsymbol{x}_{v}$ correctly classified by $\mathcal{F}_{\boldsymbol{\theta}}$. 

In adversarial settings, the objective of the adversary is to add small additive perturbation $\boldsymbol{\delta} \in \mathbb{R}^{n}$ in clean image $\boldsymbol{x}$: $\boldsymbol{x}_{adv} = \boldsymbol{x}+\boldsymbol{\delta}$. While staisfying following constraints: adversarial image follows a pertubation budget $\epsilon$ or $\norm{\boldsymbol{x_{adv}} - \boldsymbol{x}}_p \leq \epsilon$, $\boldsymbol{x}_{adv}$ looks visually similar to the true image $\boldsymbol{x}$, and the trained model generates incorrect label i.e., $\mathcal{F}_{\theta}(\boldsymbol{x}_{adv}) \neq y$. In the following sections, we assume that addition of adversarial examples changes the distribution of input from $\mathcal{P}_{\mathrm{clean}}(\boldsymbol{x},y)$ to $\mathcal{P}_{\mathrm{adv}}(\boldsymbol{x},y)$. We define adversarial accuracy as the percentage of validation examples ($\boldsymbol{x}_v$) correctly classified by a trained model $\mathcal{F}_{\theta}$ under an adversarial attack. We now consider some common methods for adding adversarial noise.

\textbf{Fast Gradient Sign Method (FGSM):}  Introduced by \citep{goodfellow2014explaining}, it exploits locally linear behaviour of neural network. It aims to generate the adversarial image $\boldsymbol{x}_{adv}$ as, 
\begin{equation*}
\boldsymbol{x}_{adv} = \boldsymbol{x}+\epsilon \cdot \text{sign} (\nabla_{\boldsymbol{x}} \mathcal{L}(\boldsymbol{x},y)).
\label{eq:FGSM}
\end{equation*}

\textbf{Basic Iterative Method (BIM):} Constructed by \cite{kurakin2016adversarial}, it is a straight forward extension of FGSM. It applies FGSM multiple times with a small step size $\alpha$ while cliping it to keep in the constraint budget. It initializes adversarial example with $\boldsymbol{x}_{adv}^{0} = \boldsymbol{x}$ and then iteratively find $x_{\mathrm{adv}}$ as,
\begin{equation*}
\centering
\boldsymbol{x}_{adv}^{N} = \text{Clip}\lbrace \boldsymbol{x}_{adv}^{N-1}+ \alpha \cdot \text{sign}(\nabla_{\boldsymbol{x}} \mathcal{L}(\boldsymbol{x}_{adv}^{N-1},y))\rbrace
\end{equation*}

where $N$ denotes iteration number for iterative attack and clip function clips all the values between 0 and 1. 

\textbf{Projected Gradient Descent (PGD):} PGD perturbs the true image $\boldsymbol{x}$ for total number of $N$ steps with smaller step sizes \cite{madry2017towards}. After each step of perturbation, PGD projects the adversarial example back onto the $\epsilon$-ball of normal image $\boldsymbol{x}$
, if it goes beyond the $\epsilon$-ball. Specifically,
\begin{equation*}
\boldsymbol{x}^{N}_{adv} = \Pi(\boldsymbol{x}^{N-1}_{adv} + \alpha \cdot \text{sign}(\nabla_{\boldsymbol{x}}\mathcal{L}(\boldsymbol{x}^{N-1}_{adv},y))),
\end{equation*}
where $\Pi$ is the projection operator, $\alpha$ is step size, and $\boldsymbol{x}^{N}_{\mathrm{adv}}$ denotes adversarial example at the $N$-th step.

\textbf{Momentum Iterative fast gradient sign Method (MIM)}.  MIM \cite{akhtar2018threat} improves the convergence of
the PGD algorithm by using the momentum. MIM generates adversarial examples by using the momentum-based iterative algorithm. By applying momentum gradient and providing techniques to escape from the poor local maximum during the iterations. The momentum gradient $\boldsymbol{g}$ can be calculated as
\begin{equation*}
    \boldsymbol{g}^{N} = d \cdot \boldsymbol{g}^{N-1} + \frac{\nabla_{\boldsymbol{x}}\mathcal{L}(\boldsymbol{x}^{N-1}_{adv},y)}{\norm{ \nabla_{\boldsymbol{x}}\mathcal{L}(\boldsymbol{x^{N-1}_{adv}},y)}  }
\end{equation*}
 where $\nabla$ shows the gradient function and $d$ is the decay factor. Initially, $\boldsymbol{x_{N-1}}$ is the original input and $\boldsymbol{g}_0$ is set to 0. In each iteration, $\boldsymbol{x}_{adv}$ is updated as
 \begin{equation*}
    \boldsymbol{x}^{N}_{adv} = \boldsymbol{x}^{N-1}_{adv} + \alpha \cdot \text{sign} (\boldsymbol{g}^{N+1} )
 \end{equation*}

\textbf{Carlini-Wagner attack (CW):} CW is an effective optimization-based attack model introduced by \cite{carlini2017towards}. It works by definining an auxilary variable $\vartheta$ and minimizes the following objective functions
\begin{equation*}
\underset{\vartheta}{\text{min}} \Vert \frac{1}{2} (\text{tanh}(\vartheta)+1) - \boldsymbol{x} \Vert + c \cdot f (\frac{1}{2}(\text{tanh}(\vartheta)+1)),
\end{equation*}
where $\frac{1}{2} (\text{tanh}(\vartheta)+1) - \boldsymbol{x}$ is the perturbation $\boldsymbol{\delta}$, c is a scalar constant, and f(.) is defined as:
\begin{equation*}
f(\boldsymbol{x}_\text{adv}) = \text{max} (\mathcal{Z}(\boldsymbol{x}_{adv})_{\boldsymbol{y}} - \text{max}\lbrace \mathcal{Z}(\boldsymbol{x}_{adv})_k : k \neq \boldsymbol{y}\rbrace, -\varrho)).
\end{equation*}
Here, $\varrho$ is to control the adversarial sample's confidence and $\mathcal{Z}_{\boldsymbol{x}_{adv}}$ are the logits values for class $k$.

 The empirical risk minimization using only clean images for training can decrease the robustness performance of DNNs. A standard approach to achieve the adversarial robustness in neural networks is adversarial training which involves fitting a neural network $\mathcal{F}_{\theta}$ on adversarially-perturbed samples \cite{kurakin2016adversarial, goodfellow2014explaining}. 
 
 We have used PGD based adversarial training as it effective against many first-order adversaries \citep{madry2017towards} unlike other methods which overfit on  single attack. Adversarial training solves following \textit{min-max optimization} problem:
\begin{equation*}
\underset{\boldsymbol{\theta}}{\text{min}}\frac{1}{N} \sum_{i=1}^{N}\underset{ \Vert\boldsymbol{\delta_{i}} \Vert \leq \epsilon}{\text{max}} \mathcal{L}(\mathcal{F}_{\boldsymbol{\theta}}(\boldsymbol{x}_i+\boldsymbol{\delta}_i),y_i)
\end{equation*}

\section{How does BatchNorm Cause Adversarial Vulnerability}
\label{sec:BN_robustness}
In this section, we explain why BatchNorm causes adversarial vulnerability. BatchNorm estimates population statistics during training by using a moving average. These estimated values are then used during inference to decrease dependence on inference examples. However, one inherent assumption of this process is that training and inference data come from same underlying distribution. Adversarial noise, on the other hand, introduces a targeted shift in the distribution of input data making this inherent assumption invalid. In the following sections, we first explain how BatchNorm works and then we empirically demonstrate our hypothesis by various experiments. 

\subsection{How BatchNorm Works}
Here, we briefly explain the working principle of BatchNorm layer which is directly realted to our hypothesis. Consider a mini-batch $\mathcal{B}$ of size $M$, containing samples $\boldsymbol{x}_i$ for $i=1,2,...,M$. BatchNorm normalizes the mini-batch by calculating the mean $\mu_{\mathcal{\beta}}$ and variance $\sigma_{\mathcal{\beta}}^2$ as follows:
\begin{align}
\centering    
  \mu_{\mathcal{B}} &= \frac{1}{M} \sum_{i=1}^{M} \boldsymbol{x}_i \quad  ;\quad
  \sigma_{\mathcal{B}} = \sqrt{\frac{1}{M} \sum_{i=1}^{M} (x_i - \mu_{\mathcal{B}})^2 + \epsilon}.
\label{eq:batch_stats}
\end{align} 
Based on these statistics, normalization is performed. To further compensate for the possible loss of representational ability of network, BatchNorm also learns per-channel linear transformation. 
\begin{equation}
\hat{\boldsymbol{x}}_i = \gamma. \frac{\boldsymbol{x}_i - \mu_{\mathcal{B}}}{\sigma_{\mathcal{B}}} + \beta
\label{eq:batch_norm}
\end{equation}

Where $\gamma$ and $\beta$ are trainable parameters that represent scale and shift, respectively. Network learns these parameters using the same optimizer - such as stochastic gradient descent - as other weights in the network. For the sake of simplicity, we will omit this linear transformation in all future discussions.

Ideally, we want to use statistics computed over all the data (population statistics) but this is not possible in mini-batch based optimization. Instead, we use expected value of mean $\mu_{\mathcal{P}} = \mathbb{E}(\mu_{\mathcal{B}})$ and variance $ \sigma^2_{\mathcal{P}} = \mathbb{E}(\sigma^2_{\mathcal{B}})$ as estimation for inference. The estimate of population statistics are computed by maintaining the moving averages of these statistics during training. Formally, moving average (also called tracking sometimes) of mean and variance are computed as follows: 

\begin{equation}
\hat{\mu}_{\mathcal{P}} = (1 - \tau)\hat{\mu}_{\mathcal{P}} + \tau \mu_{\beta}, \quad 
\hat{\sigma}_{\mathcal{P}}^{2} = (1 - \tau)\hat{\sigma}_{\mathcal{P}}^2+ \tau\sigma_{\mathcal{B}}^2
\label{eq:tracking}
\end{equation}
Here $\hat{\mu}_{\mathcal{P}}$ and $\hat{\mu}_{\mathcal{P}}$ are estimated values of population mean and variance and  $\tau$ is a hyper-parameter and weighs previous moving average and current batch statistics. For inference, BatchNorm can be represented as,
\begin{equation}
\hat{\boldsymbol{x}}_\text{test} = \frac{\boldsymbol{x}_{\text{test}} - \hat{\mu_{\mathcal{P}}}}{\hat{\sigma}_{\mathcal{P}}}
\label{eq:bn_inference}
\end{equation}

\subsection{Devil is in the Moving Average}
\begin{figure}
\centering
    \includegraphics[width=0.9\columnwidth]{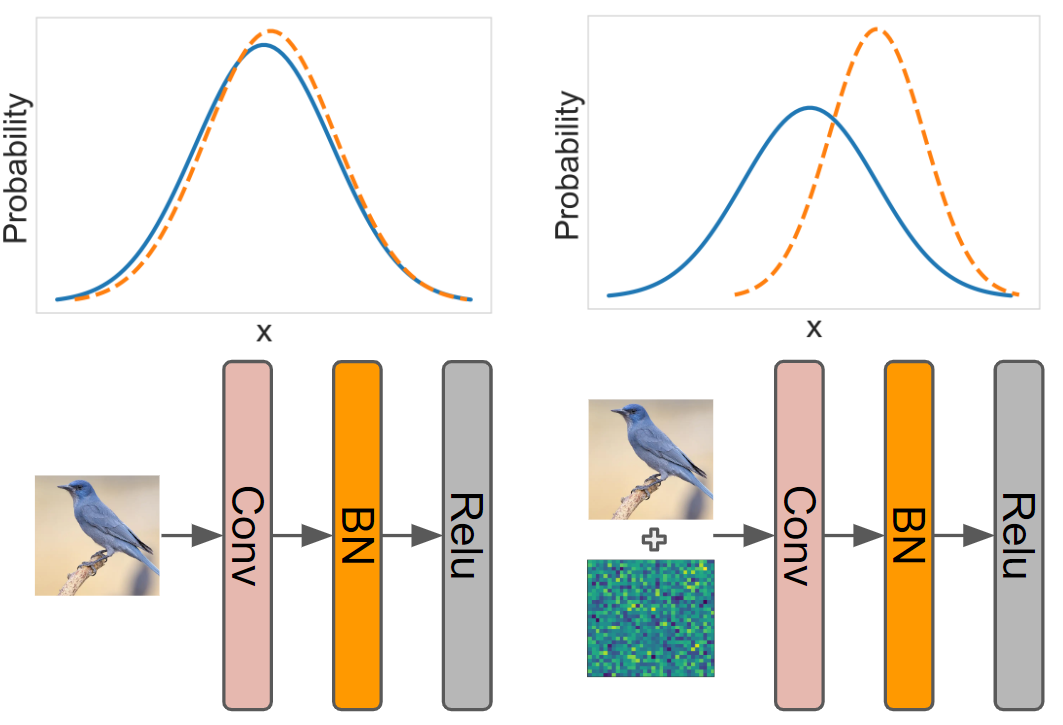}
    \caption{A conceptual illustration of the effect of adversarial distribution shift on BatchNorm. In the plot, the {\color{blue}blue line} represents an ideal distribution that BatchNorm and {\color{orange}orange line} shows the inference data distribution. Input distribution is a good approximate of ideal distribution for clean images but the distribution gets shifted when adversarial noise is added in the input image. This invalidates the implicit assumption of BatchNorm that the train and validation data will be from the same distribution. This makes population statistics estimated during training (with clean distribution) inaccurate and causes adversarial vulnerability.}
    \label{fig:distribution_shift}
\end{figure}

 At inference time, BatchNorm layer ``corrects" input with $\hat{\mu}_{\mathcal{P}}$ and $\hat{\sigma}_{\mathcal{P}}$ estimated during training. But adversarial attack introduces a targeted shift in the input. This makes estimated $\hat{\mu}_{\mathcal{P}}$ and $\hat{\sigma}_{\mathcal{P}}$ incorrect. A conceptual depiction of this is shown in Figure \ref{fig:distribution_shift}.  

\begin{figure}
    \includegraphics[width=1\columnwidth]{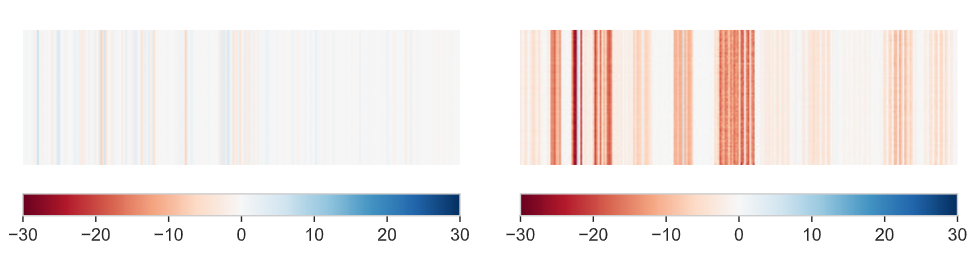}
    \caption{The difference between test batch statistics ($\mu_B$, $\sigma^2_B$) and estimated population statistics ($\hat{\mu}_{\mathcal{P}}$ and $\hat{\sigma}_{\mathcal{P}}$) under adversarial attack. The x-axis represents channels of the network and y-axis represents batches. Each line shows difference of estimated and calculated value of distribution statistics for one channel.}
    \label{fig:diff_in_tracked_batch_stats}
\end{figure}

 To show this difference, we forward propagated all the validation set samples of CIFAR10 with PGD adversarial noise added and calculated batch statistics($\mu_B$, $\sigma^2_B$) of each channel of a trained ResNet20. We then find their difference with estimated population statistics ($\hat{\mu}_{\mathcal{P}}$ and $\hat{\sigma}^2_{\mathcal{P}}$). The difference is shown in Figure \ref{fig:diff_in_tracked_batch_stats} where x-axis represents channels and the y-axis represents the difference for validation batches. The figure shows that estimated population statistics do not align with batch statistics under adversarial attack. Similarly, all the channels that are inaccurate in some batches also tend to make a similar mistake for other batches as well. Please note that the difference for one channel across batches is varying although the trend of the error is similar. The value of the difference varies for different attacks as well.

 Recent work \citep{ding2019sensitivity, jacobsen2019exploiting} has also shed light on the link of the shift in the distribution of input data and robustness. Similarly, this observation has also been used to augment adversarial examples to get SOTA results \cite{xie2019adversarial}. Based on these observations, we made the following hypothesis:
 
 \textbf{Hypothesis. } \textit{ BatchNorm's population statistics ($\hat{\mu}_{\mathcal{P}}$ and $\hat{\sigma}_{\mathcal{P}}$) are estimated from $\mathcal{P}_{\mathrm{clean}}(\boldsymbol{x},y)$ and an implicit assumption is that inference images will also come from same distribution. However, the addition of adversarial noise $\boldsymbol{\delta}$ in clean images shifts this distribution to  $\mathcal{P}_{\mathrm{adv}}(\boldsymbol{x},y)$. This breaks the assumption and hence population statistics become inaccurate. The use of these incorrect statistics makes a neural network with BatchNorm more vulnerable to adversarial inputs.}

\begin{table}[!ht]
\centering
\small
\resizebox{0.9\columnwidth}{!}{
\begin{tabular}{cc|ccccc}
\hline \hline
\textbf{Stat.} &\textbf{Normal} & \textbf{FGSM} & \textbf{BIM}   & \textbf{MIM} 
 &  \textbf{CW}&  \textbf{PGD}  \\
\hline \hline
\multicolumn{7}{c}{MNIST ($\epsilon = 0.2 $, $c=10$)} \\
\hline
P  &99.2 &59.6 &7.7 &16.7 &6.9 &07.7 \\
                       B&99.1 &\textbf{91.7} &\textbf{69.3} &\textbf{78.7} &\textbf{46.2} &\textbf{69.3}\\
 \hline  \hline

\multicolumn{7}{c}{F-MNIST ($\epsilon = 0.06 $, $c=10$)} \\
\hline
P   &93.7 &41.7 &1.5 &2.8 &02.2 &1.5\\
 B&93.6 &\textbf{73.8} &\textbf{32.4} &\textbf{41.5} &\textbf{25.8} &\textbf{32.3}\\
 \hline  \hline

\multicolumn{7}{c}{CIFAR10 ($\epsilon = 0.02 $, $c=0.01$)} \\
\hline
P &92.1 &48.3 &23.1 &27.0 &23.7 &23.1  \\ 
                       B &87.2 &\textbf{67.3} &\textbf{46.7} &\textbf{54.3} &\textbf{40.7} &\textbf{46.6} \\ 
\hline \hline

\multicolumn{7}{c}{CIFAR100 ($\epsilon = 0.02 $, $c=0.01$)} \\
\hline
P  &68.9 &20.2 &07.3 &08.5 &08.8 &07.3  \\ 
                       B &58.7 &\textbf{31.3} &\textbf{18.6} &\textbf{22.1} &\textbf{29.0} &\textbf{18.5} \\ 
\hline \hline
\multicolumn{7}{c}{ImageNet ($\epsilon = 0.008 $, $c=0.01$)} \\
\hline
P &62.7&12.5&4.0&5.2&3.4&4.4 \\ 
                       B &\textbf{63.5}&\textbf{29.2}&\textbf{18.0}&\textbf{21.4}&\textbf{21.0}&\textbf{19.0}\\   
\hline \hline
                             
\vspace{-2em}
\end{tabular} 

}

\caption{The effect of using batch statistics (B) vs estimated population statistics (P) on adversarial accuracy. Batch statistics are calculated from one batch of images (batch to be classified) from the validation set and estimated population statistics are estimated during training. The results are shown for five different datasets and five different attacks. We have used ResNet18 for Imagenet and ResNet20 for all other datasets. This table proves our hypothesis by showing an increase in accuracy if we use batch statistics that more representative of an adversarial shift in distribution.}
\label{table:tracking-robustness1}
\end{table}

According to our hypothesis, statistics calculated during training become inaccurate after an adversarial shift in the input distribution. Hence, one way to prove our hypothesis is to use adversarially perturbed validation batch to calculate these statistics and use them instead of train-time estimated statistics. Note, we only calculated mean and variance based on the input batch and no retraining of any parameter is involved. We show this for five different datasets and five different adversarial attacks in Table \ref{table:tracking-robustness1}. The clean accuracy decreases when we use BatchNorm with batch statistics (calculated from clean batch) so we expect a similar decrease in adversarial accuracy, But it instead increases. For instance, on MNIST, we get 7\% BIM adversarial accuracy with population statistics but replacing them with batch statistics from the validation set increase this to 69\%. A similar effect is also visible across all the attacks, datasets and training modes. To make our point more rigorous, we also have done experiments on different architectures with CIFAR10. The results are shown in Table \ref{tab:architectures}. A similar increase in adversarial accuracy is also visible for all of these attacks. Note, for VGG, the improvement is less than ResNet. This can have two possible explanations: VGG was designed before BatchNorm and the number of parameters in it is higher than ResNet models (models with more parameters show better adversarial accuracy \cite{madry2017towards}).

\begin{table}[ht] 
\begin{center}

\resizebox{\columnwidth}{!}{
\begin{tabular}{cc|c|ccccc}
\hline 
  \textbf{Model} &  \textbf{Stat.} & \textbf{Normal} &   \textbf{FGSM} &   \textbf{BIM} &   \textbf{MIM}&    \textbf{CW}&    \textbf{PGD} \\
\hline \hline

\multirow{2}{*}{ResNet38}&P&  93.11 &  53.62 &  25.51 &  29.51 &  27.5 &  25.52 \\
&B&  89.83 &  \textbf{74.27} &  \textbf{52.51} &  \textbf{60.72} &  \textbf{43.67} &  \textbf{52.48}\\
\hline
\multirow{2}{*}{ResNet50}&P&  93.61 &  55.97 &  29.07 &  33.45 &  29.45 &  29.07\\
 &B&  89.09 &  \textbf{72.32} &  \textbf{55.06} &  \textbf{61.61} &  \textbf{43.96} &  \textbf{55.05}\\
 \hline
\multirow{2}{*}{VGG11}&P&  91.66 &  70.21 &  63.61 &  64.71 &  57.88 &  63.61\\
 &B&  89.31 &  \textbf{81.3} &  \textbf{73.44} &  \textbf{76.79} &  \textbf{67.23} &  \textbf{73.44}\\
\hline 
\multirow{2}{*}{VGG16}&P&  93.56 &  65.58 &  53.68 &  56.28 &  48.73 &  53.68\\
  &B&  91.36 &  \textbf{81.13} &  \textbf{57.58} &  \textbf{66.32} &  \textbf{63.1} &  \textbf{57.58}\\

\hline \hline

\end{tabular} 
}
\end{center}
\caption{Table shows effect of using population vs batch statistics on adversarial accuracy for several models on CIFAR10 and Imagenet. The column Stat. shows type of statistics used for BatchNorm, P stands for Population Statistics ($\hat{\mu}_{\mathcal{P}}$, $\hat{\sigma}^2_{\mathcal{P}}$) and B stands for Batch Statistics calculated from validation batch at inference (${\mu}_{\mathcal{B}}$ and ${\sigma}^2_{\mathcal{B}}$).}
\label{tab:architectures}
\end{table}

Based on different intuitions and insights, many alternatives of BatchNorm have been introduced. Some of these variants do not require estimation of population statistics e.g. layer normalization \cite{ba2016layer}, Fixup Initialization \cite{huang2019gpipe} etc. Our hypothesis suggests that the adversarial accuracy of these variants should be higher than BatchNorm. We show results for four different alternatives in Table \ref{tab:no-tracking-norms}. Note that the clean accuracy of these alternatives is less than BatchNorm so we expect a similar drop in adversarial accuracy. On the contrary, there is an increment of adversarial accuracy, proving our hypothesis. 

\begin{table}[htp] 
\begin{center}
\resizebox{\columnwidth}{!}{
\begin{tabular}{c|c|ccccc}
\hline 
  \textbf{Norm} &  \textbf{Normal} &   \textbf{FGSM} &   \textbf{BIM} &   \textbf{MIM} &   \textbf{CW}&    \textbf{PGD} \\
\hline \hline
No Norm &82.9&43.4&29.0&31.5&28.5&29.0 \\
Fixup Init.& 91.4 & 55.7 & 37.1 & 41.1 & 17.9&39.0\\
LayerNorm &89.4 &50.8 &30.1 &33.4 &31.1 &30.1 \\
BatchNorm &92.1 &48.4 &23.1 &27.0 &23.7 &23.8 \\

\hline
\end{tabular} 
}
\end{center}
\caption{ Effects of replacing BatchNorm with alternatives that do not require any moving average therefore immune to adversarial shift. Although all of these alternatives have inferior clean accuracy; they always outperform BatchNorm in adversarial accuracy.}
\label{tab:no-tracking-norms}
\end{table}

Adversarial training leverages adversarially perturbed examples to train a neural network. An adversarially trained BatchNorm layer estimates population statistics with both clean and adversarial examples. Therefore, we should expect better adversarial accuracy which has already been shown \cite{madry2017towards}. We should also expect a smaller gap between using population statistics and input batch statistics. This indeed is true and adversarial training bridges the gap between population statistics and batch statistics based BatchNorm as shown in Table \ref{table:mvavg-robustness-adv}. For instance, on the CIFAR10 dataset, the gap between BatchNorm with batch statistics and population statistics is 100\% for regular training but it shrinks to 30\% for adversarial training. This shows the importance of the reliability of train-time estimated population statistics and their effect on the adversarial performance of a neural network. 

\begin{table}[!ht]
\small
\centering
\resizebox{0.9\columnwidth}{!}{

\begin{tabular}{c|ccccc}
\hline \hline
\textbf{Training}  & \textbf{FGSM} & \textbf{BIM}   & \textbf{MIM} 
 &  \textbf{CW}&  \textbf{PGD}  \\
 \hline \hline
\multicolumn{6}{c}{CIFAR10 } \\

\hline
Normal   &39.3  & 102.2  & 101.1  & 71.7  & 101.7 \\ 
Adversarial &23.0  & 30.5  & 31.7  & 57.3  & 30.5  \\ 
\hline \hline

\multicolumn{6}{c}{CIFAR100} \\

\hline
Normal    &55.0  & 154.8  & 160.0  & 229.5  & 153.4\\ 
Adversarial   &28.6  & 38.1  & 38.5  & 150.0  & 38.1  \\ 
\hline \hline
\end{tabular} }

\caption{Percentage adversarial accuracy gain when we use batch statistics instead of train-time estimated population statistics. Results are shown for training with clean images (normal training) and training with both clean and adversarially perturbed images (adversarial training). As expected, adversarial training shrinks the gap.}
\label{table:mvavg-robustness-adv}
\end{table}

\section{Robust Normalization}

\begin{figure}
    \centering
    \includegraphics[ width=0.6\columnwidth]{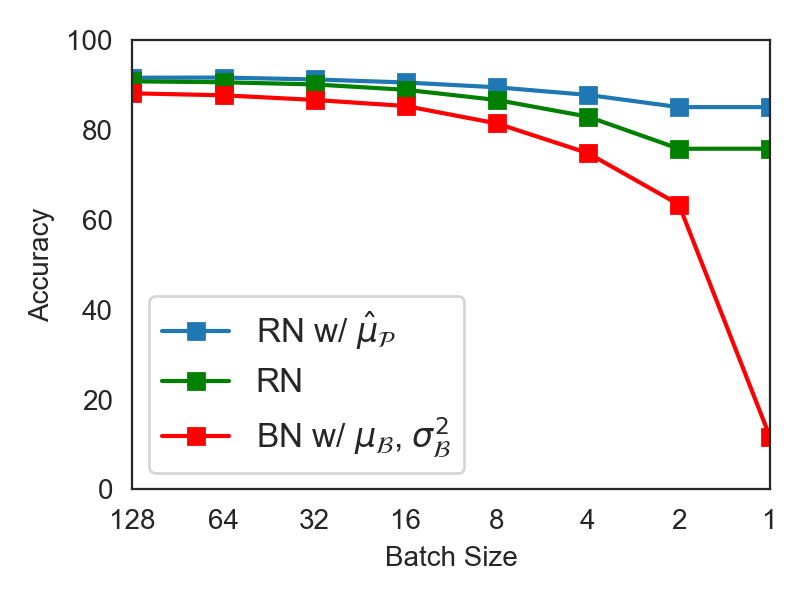}
    \vspace{-1em}
    \caption{Comparison of BatchNorm(BN) and RobustNorm(RN) for small inference batch sizes. RobustNorm performs much better when we only use inference input to compute statistics but it still suffers some loss of accuracy. We achieve performance gain by using estimated value of population mean.  }%
    \label{fig:small-bs}%
\end{figure}

In the previous section, we observed how train-time estimated population statistics in BatchNorm makes a network more vulnerable to targeted distribution shift. A straightforward solution - as shown in the experiments (Table \ref{table:tracking-robustness1}) - is to use batch statistics calculated from inference input. However, as noted by \cite{ioffe2017batch}, activations are normalized by statistics estimated from large batch during training and therefore it introduces intra-batch dependency. This makes BatchNorm dependent on moving average estimates for inference. In the experiments of the last section, we have used a batch size of 128 (same as training batch size). However, if we use small inference batch size to calculate statistics, BatchNorm performance descends to zero (see Figure \ref{fig:small-bs}). 

From recent work on understanding BatchNorm \cite{bjorck2018understanding, santurkar2018does}, we know that reduction of internal covariant shift (stability of distributions during training) is not as substantial as it was considered initially. These works highlighted the role of BatchNorm in controlling exploding activations \cite{santurkar2018does}. This leads us to the following question: can we control activations without variables that may require an estimate of population statistics? 

 We can use different data normalizations to control the activations such as normalizing with $\ell_p$ norm of the activation, rescaling between 0-1 by min-max normalization, etc. Recent work on robustness has shown a connection between the removal of outliers in activations and robustness \citep{xie2019feature, etmann2019connection}. This makes min-max normalization an ideal candidate since it rescales input (controlling exploding activations), only requires maximum and minimum values which are not dependent on the distribution (no estimates required) and can remove outliers (adversarial noise). We keep using mean considering the importance of centering the data \citep{salimans2016weight}. We define the naive version of our RobustNorm as:

\begin{equation}
\boldsymbol{y}_i = \dfrac{\boldsymbol{x}_i-\mu_{\mathcal{B}}}{r_{\mathcal{B}}}
\label{eq:minmax2}
\end{equation}

where $x_i$ is $i$-th example of batch $\mathcal{B}$, range is $r_{\mathcal{B}} = u_{\mathcal{B}} - l_{\mathcal{B}}$, maximum is $ u_{\mathcal{B}} =\underset{1\leq i \leq M}{\max}{(\boldsymbol{x}_{i})}$ and minimum is $ l_{\mathcal{B}} =\underset{1\leq i \leq M}{\min}{(\boldsymbol{x}_{i})}$.

\begin{figure}
    \centering
    \includegraphics[width=1\columnwidth]{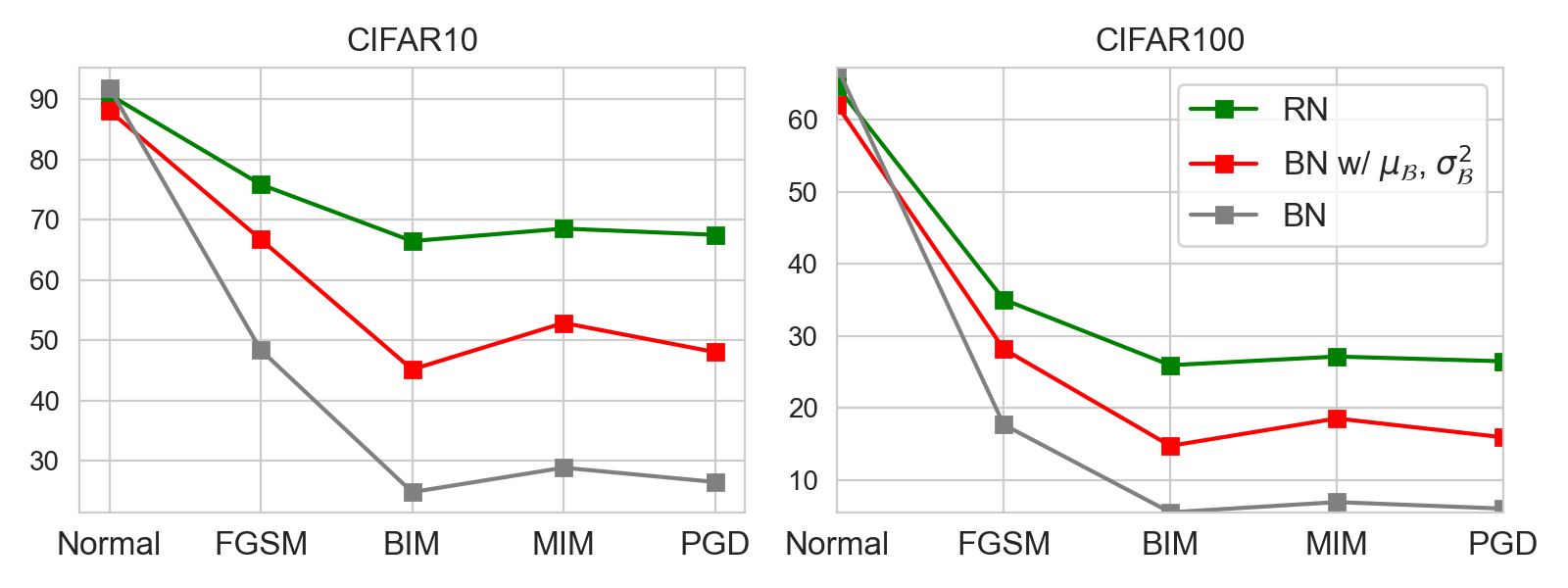}
    \vspace{-1em}
    \caption{
       Comparison of accuracy of the model with different normalization (RN: RobustNorm, BN: BatchNorm) for with ResNet20. In the presence of adversarial attacks, RobustNorm performs better than BatchNorm. }%
    \label{fig:rn_adv_robustness}%
\end{figure}

From von Szokefalvi Nagy inequality ($r_{\mathcal{B}}^2 \leq 2n \sigma_{\mathcal{B}}^2$, where $n$ is number of samples to estimate range), we can say that range supresses activations with higher intensity than the variance. BatchNorm uses linear transform to project activations to an appropriate range. However, in our case, range makes it harder to learn this projection in the start of learning. To make the control more flexible, we introduced a new hyper-parameter -- norm power ($p$). Finally, we define Robust Normalization (RobustNorm or RN) as follows:
\begin{equation}
\boldsymbol{y}_i = \dfrac{\boldsymbol{x}_i-\mu_{\mathcal{B}}}{r_{\mathcal{B}}^p}.
\label{eq:rboust_norm}
\end{equation}
 We only fine tune this $p$ for better convergence and generlizability across datasets (for details, see Section \ref{sec:power_hyperparameter}). We evuluated robustness of RobustNorm for three different datasets. RobustNorm's accuracy is higher in the presence of adversarial attacks (Figure \ref{fig:rn_adv_robustness}). Specifically, RobustNorm increases adversarial accuracy of ResNet20 from 22\% to 70\% for CIFAR10. All the results are shown in Figure \ref{fig:rn_adv_robustness}. 

\begin{figure}
    \centering
    \includegraphics[width=1\columnwidth]{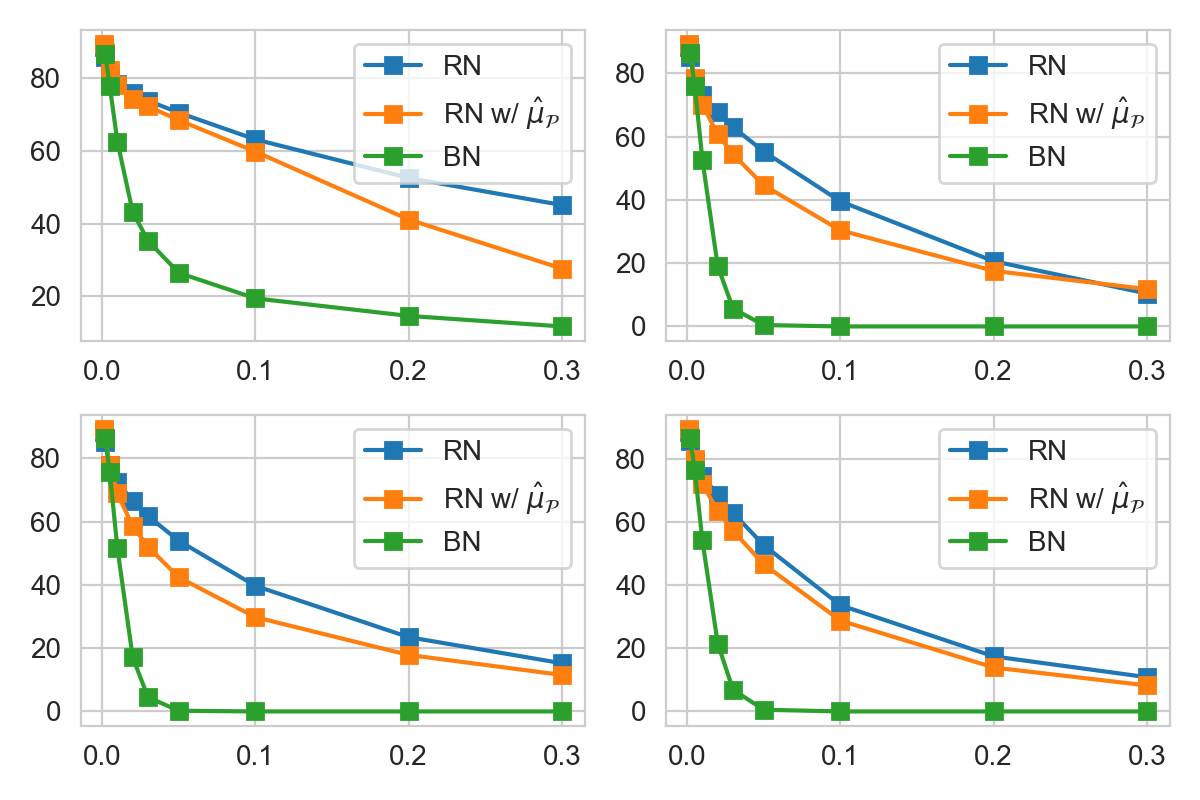}
    \caption{Effect of increasing $\epsilon$ for different normalizations. We compare adversarial accuracy in the presence of FGSM (upper left), BIM (upper right), MIM (lower left) and PGD (lower right) for RobustNorm and BatchNorm.  BN: BatchNorm, RN: RobustNorm and RN w/ $\hat{\mu}_{\mathcal{P}}$: RobustNorm with the population mean. RobustNorm consistently performs well in the presence of increasing adversarial noise.}%
    \label{fig:adv_robustness}%
\end{figure}

RobustNorm performs better compared to BatchNorm when we only use inference inputs to calculate statistics as shown in Figure \ref{fig:small-bs}. But, it still suffers some loss of accuracy. Since mean ($\mu$) is a distribution statistics, we use its estimate calculated during training. This improved the performance of RobustNorm for small batch size is shown in Figure \ref{fig:small-bs}. To understand the effect of $\hat{\mu}_{\mathcal{P}}$ on adversarial accuracy of RobustNorm, we perform experiments with varying values of $\epsilon$. As shown in Figure \ref{fig:adv_robustness}, adversarial accuracy of RobustNorm with $\hat{\mu}_{\mathcal{P}}$ is comparable to RobustNorm while also having consistent small inference batch performance.

\section{Ablation Studies}
In this section, we have validated and explored different properties of RobustNorm. 

\subsection{Experimental Setup}
\label{sec:experimental_setup}
We have used two network architectures, ResNet \cite{he2016deep} with 20,38 and 50 layers and VGG \cite{simonyan2014very} with 11 and 16 layers. We choose ResNet and VGG because they represent two diverse families of architectures, and can be considered as a baseline of many of the networks commonly used in deep learning. In ResNet family, we have DenseNet \cite{huang2017densely}, WideResNet \cite{zagoruyko2016wide}, ResNext \cite{xie2017aggregated}, ResNet with Stochastic depth \cite{huang2016deep} etc., and VGG can be related to LeNet \cite{lecun1998gradient}, AlexNet \cite{krizhevsky2012imagenet} etc. We choose to use ResNet20 as our baseline architecture along with the CIFAR10 dataset because of the ease of training and quick experimentation with limited available compute power. However, we also presented results on all of the other networks and datasets to show our point more rigorously whenever possible. We have also used five different datasets: MNIST \cite{lecun-mnisthandwrittendigit-2010}, Fashion-MNIST \cite{xiao2017fashion}, CIFAR10, CIFAR100 \cite{krizhevsky2009learning} and Imagenet \cite{deng2009imagenet}. We have always used a learning rate of 0.1 except for no normalization scenarios where convergence is not possible with higher learning rates. In such cases, we have used a learning rate of 0.01. We decrease the learning rate 10 times at 80th and 120th epoch for CIFAR10, 100; at 30th epoch for MNIST, Fashion-MNIST; and at 30th, 60th and 90th epoch for Imagenet. We trained CIFAR10, 100 for 164 epochs, MNIST, Fashion-MNIST for 50 epochs and ImageNet for 100 epochs. We used \cite{madry2017towards}'s standard settings for adversarial training. We have tried to stick with standard training procedures as much as possible. For all the attacks where distance measure is required, we use $\ell_{\inf}$-norm as it is more difficult to evade.

\subsection{Analysis of Power Hyperparameter}
\label{sec:power_hyperparameter}
\begin{figure}
    \centering
    \includegraphics[width=1\columnwidth]{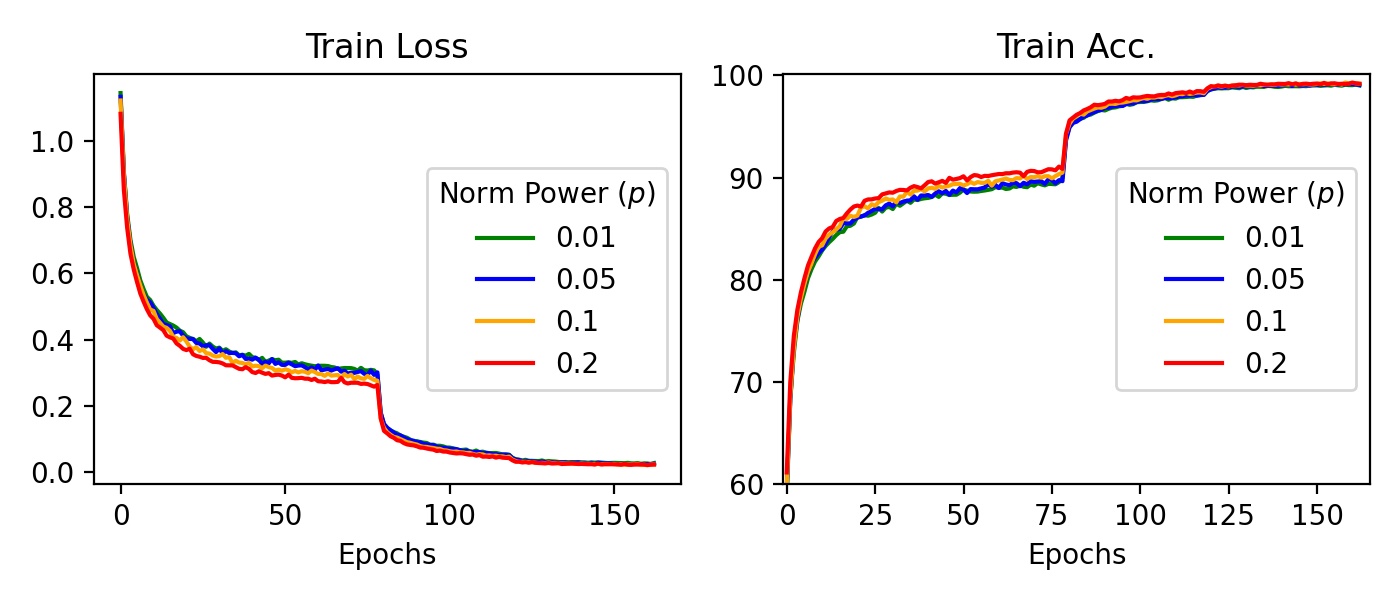}
    \caption{Training curves on CIFAR10 dataset for RobustNorm. We compare loss (left) and acuracy (right) of RobustNorm with different hyperparameter power ($p$) values. Note that $p=0.2$ (red line) converges faster than other.}
    \label{fig:convergence}
\end{figure}

\begin{figure}
    \centering
    \includegraphics[width=1\columnwidth]{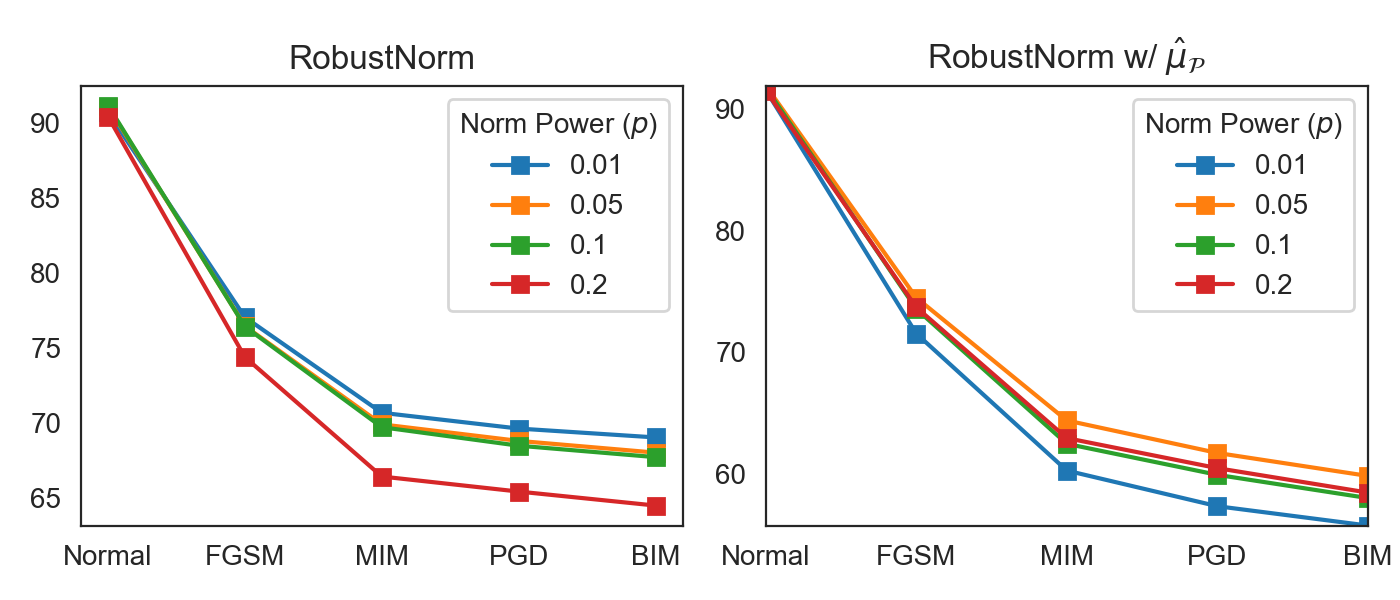}
    \caption{Effect of hyperparameter $p$ on adversaril robustness. The results are shwon for CIFAR10 and ResNet20. This shows room for improvement by tuning it.}
    \label{fig:power}
\end{figure}

The RobustNorm introduces a new hyperparameter called power or $p$ of the range $r_{\mathcal{B}}$. We found $p=0.2$ having faster convergence (see Figure (\ref{fig:convergence}), red shows $p=0.2$) and generality across datasets. Therefore, we have used it for all our experiments. Later, we observed that faster convergence does not necessarily mean better adversarial robustness (see Figure \ref{fig:power}). For instance, RobustNorm with $p=0.2$ performs worse in terms of adversarial accuracy when compared to other values. Similarly, $p=0.05$ has better adversarial robustness in RobustNorm with using population mean. This shows room for more improvement by tuning this hyperparameter. 


\subsection{Effect of Adversarial Attack Budget $\epsilon$}
Another important aspect of the robustness of a network is how it responds to the increasing value of adversarial noise ($\epsilon$). To understand it, we evaluated RobustNorm and RobustNorm with the estimated population mean for an increasing value of $p$. RobustNorm performs significantly better compared to BatchNorm. For instance, the performance of ResNet20 with BatchNorm sinks to zero for $\epsilon>0.05$ for BIM or PGD attack. RobustNorm, on the other hand, is more resilient to even higher $\epsilon$ and consistently performs better compared to BatchNorm. RobustNorm with the population mean perform better than BatchNorm although we lose some adversarial performance gain compare to RobustNorm.

\subsection{Scalibility to Different Architectures}
We also evaluate RobustNorm to show its scalability on different neural network architectures and depths. We choose ResNet and VGG architectures as a wide variety of neural networks evolved from these networks. Similarly, VGG was designed before BatchNorm so it is also interesting to see its performance under different normalizations. To show the scalability of RobustNorm for different depths, we choose two commonly used depths of ResNet (38 and 50) and VGG (11 and 16). Results for the experiments on these architectures for CIFAR10 are shown in Table \ref{tab:architecturesRN}. RobustNorm outperforms BatchNorm by wide margins in all of these networks. For instance, RobustNorm has a margin of 50\% with ResNet38, 31\% for ResNet50, 15\% for VGG11 and 28\% for VGG16 when the input has BIM adversarial noise. Similar trends are also visible under different attacks.
\begin{table}[ht]
    \centering
\resizebox{\columnwidth}{!}{

\begin{tabular}{llrrrrr}
\toprule
    Model &   Norm &   FGSM &  BIM &    MIM &   CW &  PGD \\
\midrule
\multirow{3}{*}{ResNet38}&     BN &  49.10 &    18.77 &  23.90 &  12.74 &    20.80 \\
                         &     RN &  78.71 &    71.19 &  72.74 &  29.17 &    72.28 \\
                         &     RN w/ $\hat{\mu}_{\mathcal{P}}$  &  80.29 &    71.13 &  73.23 &  31.02 &    72.46 \\
 \hline
\multirow{3}{*}{ResNet50}&     BN &  51.14 &    24.31 &  29.20 &  12.53 &    26.29 \\
                         &     RN &  76.58 &    65.97 &  68.82 &  32.19 &    67.53 \\
                         &    RN w/ $\hat{\mu}_{\mathcal{P}}$  &  80.10 &    71.02 &  72.64 &  35.76 &    72.07 \\
 \hline
\multirow{3}{*}{VGG11}&    BN &  69.44 &    61.83 &  63.70 &  42.88 &    62.79 \\
                         &   RN &  82.03 &    77.76 &  78.94 &  61.57 &    78.24 \\
                         &  RN w/ $\hat{\mu}_{\mathcal{P}}$ &  82.05 &    76.11 &  77.88 &  59.47 &    76.95 \\
    \hline
\multirow{3}{*}{VGG16}&     BN &  63.60 &    50.15 &  53.53 &  30.13 &    51.52 \\
                         &     RN &  83.56 &    78.75 &  80.12 &  50.10 &    79.28 \\
                         &    RN w/ $\hat{\mu}_{\mathcal{P}}$  &  82.74 &    74.73 &  76.99 &  49.57 &    75.81 \\
\bottomrule
\end{tabular} 
}

    \caption{Scalibility of RobustNorm across different architectures and depths. BN: BatchNorm, RN: RobustNorm and RN w/ $\hat{\mu}_{\mathcal{P}}$: RobustNorm with the population mean. Note consistent performance of RobustNorm across architectures and depths.  }
    \label{tab:architecturesRN}
\end{table}
\subsection{RobustNorm for ImageNet}
It is well known that adversarial defense methods are difficult to scale on large datasets such as ImageNet \cite{kurakin2016adversarial}. To test the effectiveness of RobustNorm at scale, we performed experiments for RobustNorm on ImageNet. Results are shown in Table \ref{tab:imagenet}. RobustNorm beats BatchNorm for all the attacks by a wide margin. Note that we have not used any fine-tuning for hyper-parameter $p$ due to limited compute power available. 
\begin{table}[ht]
    \centering
\resizebox{\columnwidth}{!}{

\begin{tabular}{llrrrrr}
\toprule
   Norm& Normal &   FGSM &  BIM &    MIM &   CW &  PGD \\
\midrule
BN &62.9&12.5&4.0&5.2&3.4&4.4 \\
RN &61.6 &28.5 &19.8 &21.9 &22.8 &20.5 \\
RN w/ $\hat{\mu}_{\mathcal{P}}$&61.8 &28.1 &17.7 &20.1 &25.3 &18.7 \\

\hline
\bottomrule
\end{tabular} 
}

    \caption{Comparison of RobustNorm (RN) and BatchNorm (BN) for ImageNet. We have used ResNet18 and $\epsilon=0.008$. RobustNorm perform better than BatchNorm for all the attacks. }
    \label{tab:imagenet}
\end{table}

\section{Limitations and Future Work}
 We believe that several factors can help fully understand the role of BatchNorm in the adversarial vulnerability of a network such as the effect of intra-batch dependency of BatchNorm on adversarial behavior as well its poor performance for small batches. Similarly, input image at inference time is not used to ``correct" population statistics. As a better estimate of statistics is essential, this correction can help make the neural network more robust. Another exciting direction is to see the generalizability of RobustNorm to Layer, Group, and InstanceNorm. Since these norms do not require an estimate of population statistics, RobustNorm can work better.

\section{Conclusion}

We have investigated how BatchNorm makes a network more vulnerable to adversarial attacks. We observed that BatchNorm estimates population statistics from clean images during training and the addition of adversarial noise introduces a targeted distribution shift in the input. We hypothesized that this shift makes train-time estimated statistics inaccurate thereby causing the adversarial vulnerability. We showed our hypothesis by showing adversarial accuracy differences between statistics calculated form input batch and train-time estimated statistics. The results on multiple datasets and architectures proved our hypothesis. We also showed that normalizations that do not require these train-time estimated values perform better compare to BatchNorm. Afterward, based on these insights and intuitions from previous work understanding of BatchNorm, we proposed a new normalization called  RobustNorm. RobustNorm offers a promise of improving the robustness of any model with BatchNorm while having other benefits. We demonstrate this on ResNet and VGG for different depths.

\bibliography{refs}

\begin{thebibliography}{46}
\providecommand{\natexlab}[1]{#1}
\providecommand{\url}[1]{\texttt{#1}}
\expandafter\ifx\csname urlstyle\endcsname\relax
  \providecommand{\doi}[1]{doi: #1}\else
  \providecommand{\doi}{doi: \begingroup \urlstyle{rm}\Url}\fi

\bibitem[Akhtar \& Mian(2018)Akhtar and Mian]{akhtar2018threat}
Akhtar, N. and Mian, A.
\newblock Threat of adversarial attacks on deep learning in computer vision: A
  survey.
\newblock \emph{IEEE Access}, 6:\penalty0 14410--14430, 2018.

\bibitem[Ba et~al.(2016)Ba, Kiros, and Hinton]{ba2016layer}
Ba, J.~L., Kiros, J.~R., and Hinton, G.~E.
\newblock Layer normalization.
\newblock \emph{arXiv preprint arXiv:1607.06450}, 2016.

\bibitem[Bjorck et~al.(2018)Bjorck, Gomes, Selman, and
  Weinberger]{bjorck2018understanding}
Bjorck, N., Gomes, C.~P., Selman, B., and Weinberger, K.~Q.
\newblock Understanding batch normalization.
\newblock In \emph{Advances in Neural Information Processing Systems}, pp.\
  7694--7705, 2018.

\bibitem[Carlini \& Wagner(2017)Carlini and Wagner]{carlini2017towards}
Carlini, N. and Wagner, D.
\newblock Towards evaluating the robustness of neural networks.
\newblock In \emph{2017 IEEE Symposium on Security and Privacy (SP)}, pp.\
  39--57. IEEE, 2017.

\bibitem[Deng et~al.(2009)Deng, Dong, Socher, Li, Li, and
  Fei-Fei]{deng2009imagenet}
Deng, J., Dong, W., Socher, R., Li, L.-J., Li, K., and Fei-Fei, L.
\newblock Imagenet: A large-scale hierarchical image database.
\newblock In \emph{2009 IEEE conference on computer vision and pattern
  recognition}, pp.\  248--255. Ieee, 2009.

\bibitem[Ding et~al.(2019)Ding, Lui, Jin, Wang, and Huang]{ding2019sensitivity}
Ding, G.~W., Lui, K. Y.~C., Jin, X., Wang, L., and Huang, R.
\newblock On the sensitivity of adversarial robustness to input data
  distributions.
\newblock 2019.

\bibitem[Etmann et~al.(2019)Etmann, Lunz, Maass, and
  Sch{\"o}nlieb]{etmann2019connection}
Etmann, C., Lunz, S., Maass, P., and Sch{\"o}nlieb, C.-B.
\newblock On the connection between adversarial robustness and saliency map
  interpretability.
\newblock \emph{arXiv preprint arXiv:1905.04172}, 2019.

\bibitem[Fawzi et~al.(2016)Fawzi, Moosavi-Dezfooli, and
  Frossard]{fawzi2016robustness}
Fawzi, A., Moosavi-Dezfooli, S.-M., and Frossard, P.
\newblock Robustness of classifiers: from adversarial to random noise.
\newblock In \emph{Advances in Neural Information Processing Systems}, pp.\
  1632--1640, 2016.

\bibitem[Fawzi et~al.(2018)Fawzi, Fawzi, and Fawzi]{fawzi2018adversarial}
Fawzi, A., Fawzi, H., and Fawzi, O.
\newblock Adversarial vulnerability for any classifier.
\newblock In \emph{Advances in Neural Information Processing Systems}, pp.\
  1178--1187, 2018.

\bibitem[Ford et~al.(2019)Ford, Gilmer, Carlini, and
  Cubuk]{ford2019adversarial}
Ford, N., Gilmer, J., Carlini, N., and Cubuk, D.
\newblock Adversarial examples are a natural consequence of test error in
  noise.
\newblock \emph{arXiv preprint arXiv:1901.10513}, 2019.

\bibitem[Galloway et~al.(2019)Galloway, Golubeva, Tanay, Moussa, and
  Taylor]{galloway2019batch}
Galloway, A., Golubeva, A., Tanay, T., Moussa, M., and Taylor, G.~W.
\newblock Batch normalization is a cause of adversarial vulnerability.
\newblock \emph{arXiv preprint arXiv:1905.02161}, 2019.

\bibitem[Geirhos et~al.(2018)Geirhos, Rubisch, Michaelis, Bethge, Wichmann, and
  Brendel]{geirhos2018imagenet}
Geirhos, R., Rubisch, P., Michaelis, C., Bethge, M., Wichmann, F.~A., and
  Brendel, W.
\newblock Imagenet-trained cnns are biased towards texture; increasing shape
  bias improves accuracy and robustness.
\newblock \emph{arXiv preprint arXiv:1811.12231}, 2018.

\bibitem[Gilmer et~al.(2018)Gilmer, Metz, Faghri, Schoenholz, Raghu,
  Wattenberg, and Goodfellow]{gilmer2018adversarial}
Gilmer, J., Metz, L., Faghri, F., Schoenholz, S.~S., Raghu, M., Wattenberg, M.,
  and Goodfellow, I.
\newblock Adversarial spheres.
\newblock \emph{arXiv preprint arXiv:1801.02774}, 2018.

\bibitem[Goodfellow et~al.(2014)Goodfellow, Shlens, and
  Szegedy]{goodfellow2014explaining}
Goodfellow, I.~J., Shlens, J., and Szegedy, C.
\newblock Explaining and harnessing adversarial examples.
\newblock \emph{arXiv preprint arXiv:1412.6572}, 2014.

\bibitem[He et~al.(2016)He, Zhang, Ren, and Sun]{he2016deep}
He, K., Zhang, X., Ren, S., and Sun, J.
\newblock Deep residual learning for image recognition.
\newblock In \emph{Proceedings of the IEEE conference on computer vision and
  pattern recognition}, pp.\  770--778, 2016.

\bibitem[Hoffer et~al.(2018)Hoffer, Banner, Golan, and Soudry]{hoffer2018norm}
Hoffer, E., Banner, R., Golan, I., and Soudry, D.
\newblock Norm matters: efficient and accurate normalization schemes in deep
  networks.
\newblock In \emph{Advances in Neural Information Processing Systems}, pp.\
  2160--2170, 2018.

\bibitem[Huang et~al.(2016)Huang, Sun, Liu, Sedra, and
  Weinberger]{huang2016deep}
Huang, G., Sun, Y., Liu, Z., Sedra, D., and Weinberger, K.~Q.
\newblock Deep networks with stochastic depth.
\newblock In \emph{European conference on computer vision}, pp.\  646--661.
  Springer, 2016.

\bibitem[Huang et~al.(2017)Huang, Liu, Van Der~Maaten, and
  Weinberger]{huang2017densely}
Huang, G., Liu, Z., Van Der~Maaten, L., and Weinberger, K.~Q.
\newblock Densely connected convolutional networks.
\newblock In \emph{Proceedings of the IEEE conference on computer vision and
  pattern recognition}, pp.\  4700--4708, 2017.

\bibitem[Huang et~al.(2019)Huang, Cheng, Bapna, Firat, Chen, Chen, Lee, Ngiam,
  Le, Wu, et~al.]{huang2019gpipe}
Huang, Y., Cheng, Y., Bapna, A., Firat, O., Chen, D., Chen, M., Lee, H., Ngiam,
  J., Le, Q.~V., Wu, Y., et~al.
\newblock Gpipe: Efficient training of giant neural networks using pipeline
  parallelism.
\newblock In \emph{Advances in Neural Information Processing Systems}, pp.\
  103--112, 2019.

\bibitem[Ilyas et~al.(2019)Ilyas, Santurkar, Tsipras, Engstrom, Tran, and
  Madry]{ilyas2019adversarial}
Ilyas, A., Santurkar, S., Tsipras, D., Engstrom, L., Tran, B., and Madry, A.
\newblock Adversarial examples are not bugs, they are features.
\newblock In \emph{Advances in Neural Information Processing Systems}, pp.\
  125--136, 2019.

\bibitem[Ioffe(2017)]{ioffe2017batch}
Ioffe, S.
\newblock Batch renormalization: Towards reducing minibatch dependence in
  batch-normalized models.
\newblock In \emph{Advances in neural information processing systems}, pp.\
  1945--1953, 2017.

\bibitem[Jacobsen et~al.(2018)Jacobsen, Behrmann, Zemel, and
  Bethge]{jacobsen2018excessive}
Jacobsen, J.-H., Behrmann, J., Zemel, R., and Bethge, M.
\newblock Excessive invariance causes adversarial vulnerability.
\newblock \emph{arXiv preprint arXiv:1811.00401}, 2018.

\bibitem[Jacobsen et~al.(2019)Jacobsen, Behrmannn, Carlini, Tramer, and
  Papernot]{jacobsen2019exploiting}
Jacobsen, J.-H., Behrmannn, J., Carlini, N., Tramer, F., and Papernot, N.
\newblock Exploiting excessive invariance caused by norm-bounded adversarial
  robustness.
\newblock \emph{arXiv preprint arXiv:1903.10484}, 2019.

\bibitem[Krizhevsky et~al.(2012)Krizhevsky, Sutskever, and
  Hinton]{krizhevsky2012imagenet}
Krizhevsky, A., Sutskever, I., and Hinton, G.~E.
\newblock Imagenet classification with deep convolutional neural networks.
\newblock In \emph{Advances in neural information processing systems}, pp.\
  1097--1105, 2012.

\bibitem[Krizhevsky et~al.(2009)]{krizhevsky2009learning}
Krizhevsky, A. et~al.
\newblock Learning multiple layers of features from tiny images.
\newblock Technical report, Citeseer, 2009.

\bibitem[Kurakin et~al.(2016)Kurakin, Goodfellow, and
  Bengio]{kurakin2016adversarial}
Kurakin, A., Goodfellow, I., and Bengio, S.
\newblock Adversarial machine learning at scale.
\newblock \emph{arXiv preprint arXiv:1611.01236}, 2016.

\bibitem[LeCun \& Cortes(2010)LeCun and
  Cortes]{lecun-mnisthandwrittendigit-2010}
LeCun, Y. and Cortes, C.
\newblock {MNIST} handwritten digit database.
\newblock 2010.
\newblock URL \url{http://yann.lecun.com/exdb/mnist/}.

\bibitem[LeCun et~al.(1998)LeCun, Bottou, Bengio, and
  Haffner]{lecun1998gradient}
LeCun, Y., Bottou, L., Bengio, Y., and Haffner, P.
\newblock Gradient-based learning applied to document recognition.
\newblock \emph{Proceedings of the IEEE}, 86\penalty0 (11):\penalty0
  2278--2324, 1998.

\bibitem[Madry et~al.(2017)Madry, Makelov, Schmidt, Tsipras, and
  Vladu]{madry2017towards}
Madry, A., Makelov, A., Schmidt, L., Tsipras, D., and Vladu, A.
\newblock Towards deep learning models resistant to adversarial attacks.
\newblock \emph{arXiv preprint arXiv:1706.06083}, 2017.

\bibitem[Nguyen et~al.(2015)Nguyen, Yosinski, and Clune]{nguyen2015deep}
Nguyen, A., Yosinski, J., and Clune, J.
\newblock Deep neural networks are easily fooled: High confidence predictions
  for unrecognizable images.
\newblock In \emph{Proceedings of the IEEE conference on computer vision and
  pattern recognition}, pp.\  427--436, 2015.

\bibitem[Papernot et~al.(2016)Papernot, McDaniel, Wu, Jha, and
  Swami]{papernot2016distillation}
Papernot, N., McDaniel, P., Wu, X., Jha, S., and Swami, A.
\newblock Distillation as a defense to adversarial perturbations against deep
  neural networks.
\newblock In \emph{2016 IEEE Symposium on Security and Privacy (SP)}, pp.\
  582--597. IEEE, 2016.

\bibitem[Salimans \& Kingma(2016)Salimans and Kingma]{salimans2016weight}
Salimans, T. and Kingma, D.~P.
\newblock Weight normalization: A simple reparameterization to accelerate
  training of deep neural networks.
\newblock In \emph{Advances in Neural Information Processing Systems}, pp.\
  901--909, 2016.

\bibitem[Santurkar et~al.(2018)Santurkar, Tsipras, Ilyas, and
  Madry]{santurkar2018does}
Santurkar, S., Tsipras, D., Ilyas, A., and Madry, A.
\newblock How does batch normalization help optimization?
\newblock In \emph{Advances in Neural Information Processing Systems}, pp.\
  2483--2493, 2018.

\bibitem[Schmidt et~al.(2018)Schmidt, Santurkar, Tsipras, Talwar, and
  Madry]{schmidt2018adversarially}
Schmidt, L., Santurkar, S., Tsipras, D., Talwar, K., and Madry, A.
\newblock Adversarially robust generalization requires more data.
\newblock In \emph{Advances in Neural Information Processing Systems}, pp.\
  5014--5026, 2018.

\bibitem[Simon-Gabriel et~al.(2018)Simon-Gabriel, Ollivier, Bottou,
  Sch{\"o}lkopf, and Lopez-Paz]{simon2018adversarial}
Simon-Gabriel, C.-J., Ollivier, Y., Bottou, L., Sch{\"o}lkopf, B., and
  Lopez-Paz, D.
\newblock Adversarial vulnerability of neural networks increases with input
  dimension.
\newblock \emph{arXiv preprint arXiv:1802.01421}, 2018.

\bibitem[Simonyan \& Zisserman(2014)Simonyan and Zisserman]{simonyan2014very}
Simonyan, K. and Zisserman, A.
\newblock Very deep convolutional networks for large-scale image recognition.
\newblock \emph{arXiv preprint arXiv:1409.1556}, 2014.

\bibitem[Szegedy et~al.(2013)Szegedy, Zaremba, Sutskever, Bruna, Erhan,
  Goodfellow, and Fergus]{szegedy2013intriguing}
Szegedy, C., Zaremba, W., Sutskever, I., Bruna, J., Erhan, D., Goodfellow, I.,
  and Fergus, R.
\newblock Intriguing properties of neural networks.
\newblock \emph{arXiv preprint arXiv:1312.6199}, 2013.

\bibitem[Tram{\`e}r et~al.(2017)Tram{\`e}r, Kurakin, Papernot, Goodfellow,
  Boneh, and McDaniel]{tramer2017ensemble}
Tram{\`e}r, F., Kurakin, A., Papernot, N., Goodfellow, I., Boneh, D., and
  McDaniel, P.
\newblock Ensemble adversarial training: Attacks and defenses.
\newblock \emph{arXiv preprint arXiv:1705.07204}, 2017.

\bibitem[Ulyanov et~al.(2016)Ulyanov, Vedaldi, and
  Lempitsky]{ulyanov2016instance}
Ulyanov, D., Vedaldi, A., and Lempitsky, V.
\newblock Instance normalization: The missing ingredient for fast stylization.
\newblock \emph{arXiv preprint arXiv:1607.08022}, 2016.

\bibitem[Wu \& He(2018)Wu and He]{wu2018group}
Wu, Y. and He, K.
\newblock Group normalization.
\newblock In \emph{Proceedings of the European Conference on Computer Vision
  (ECCV)}, pp.\  3--19, 2018.

\bibitem[Xiao et~al.(2017)Xiao, Rasul, and Vollgraf]{xiao2017fashion}
Xiao, H., Rasul, K., and Vollgraf, R.
\newblock Fashion-mnist: a novel image dataset for benchmarking machine
  learning algorithms.
\newblock \emph{arXiv preprint arXiv:1708.07747}, 2017.

\bibitem[Xie et~al.(2019{\natexlab{a}})Xie, Tan, Gong, Wang, Yuille, and
  Le]{xie2019adversarial}
Xie, C., Tan, M., Gong, B., Wang, J., Yuille, A., and Le, Q.~V.
\newblock Adversarial examples improve image recognition.
\newblock \emph{arXiv preprint arXiv:1911.09665}, 2019{\natexlab{a}}.

\bibitem[Xie et~al.(2019{\natexlab{b}})Xie, Wu, Maaten, Yuille, and
  He]{xie2019feature}
Xie, C., Wu, Y., Maaten, L. v.~d., Yuille, A.~L., and He, K.
\newblock Feature denoising for improving adversarial robustness.
\newblock In \emph{Proceedings of the IEEE Conference on Computer Vision and
  Pattern Recognition}, pp.\  501--509, 2019{\natexlab{b}}.

\bibitem[Xie et~al.(2017)Xie, Girshick, Doll{\'a}r, Tu, and
  He]{xie2017aggregated}
Xie, S., Girshick, R., Doll{\'a}r, P., Tu, Z., and He, K.
\newblock Aggregated residual transformations for deep neural networks.
\newblock In \emph{Proceedings of the IEEE conference on computer vision and
  pattern recognition}, pp.\  1492--1500, 2017.

\bibitem[Yuan et~al.(2019)Yuan, He, Zhu, and Li]{yuan2019adversarial}
Yuan, X., He, P., Zhu, Q., and Li, X.
\newblock Adversarial examples: Attacks and defenses for deep learning.
\newblock \emph{IEEE transactions on neural networks and learning systems},
  2019.

\bibitem[Zagoruyko \& Komodakis(2016)Zagoruyko and
  Komodakis]{zagoruyko2016wide}
Zagoruyko, S. and Komodakis, N.
\newblock Wide residual networks.
\newblock \emph{arXiv preprint arXiv:1605.07146}, 2016.

\end{thebibliography}
\bibliographystyle{icml2019}

\end{document}